\documentclass[10pt,twocolumn,letterpaper]{article}

\usepackage{iccv}
\usepackage{times}
\usepackage{epsfig}
\usepackage{graphicx}
\usepackage{amsmath}
\usepackage{amssymb}

\usepackage{multirow}
\usepackage{makecell}
\usepackage{booktabs}
\usepackage{xurl}

\usepackage[pagebackref=true,breaklinks=true,letterpaper=true,colorlinks,bookmarks=false]{hyperref}

\iccvfinalcopy 

\def\iccvPaperID{8103} 

\makeatletter
\def\thanks#1{\protected@xdef\@thanks{\@thanks
        \protect\footnotetext{#1}}}
\makeatother

\newcommand\tb[1]{\textbf{#1}}

\newcommand{\bd}[1]{\textcolor[rgb]{ 0,0,0}{#1}}

\newcommand{\jq}[1]{\textcolor[rgb]{0,0,0}{#1}}
\ificcvfinal\fi

\begin{document}

\title{Object Tracking by Jointly Exploiting Frame and Event Domain}

\author{Jiqing Zhang$^{1,*}$, Xin Yang$^{1,*}$, Yingkai Fu$^1$, Xiaopeng Wei$^1$, Baocai Yin$^{1,\dag}$, Bo Dong$^{2,\dag}$\\
$^1$Dalian University of Technology, $^2$ SRI International \thanks{$^*$ Joint first authors.
$^\dag$Baocai Yin (ybc@dlut.edu.cn) and Bo Dong (bo.dong@sri.com) are the corresponding authors.} \\}

\maketitle
\ificcvfinal\thispagestyle{empty}\fi

\begin{abstract}
\bd{Inspired by the complementarity between conventional frame-based and bio-inspired event-based cameras, we propose a multi-modal based approach to fuse visual cues from the frame- and event-domain to enhance the single object tracking performance, especially in degraded conditions (\eg, scenes with high dynamic range, low light, and fast-motion objects). The proposed approach can effectively and adaptively combine meaningful information from both domains. Our approach's effectiveness is enforced by a novel designed cross-domain attention schemes, which can effectively enhance features based on self- and cross-domain attention schemes; The adaptiveness is guarded by a specially designed weighting scheme, which can adaptively balance the contribution of the two domains. To exploit event-based visual cues in single-object tracking, we construct a large-scale frame-event-based dataset, which we subsequently employ to train a novel frame-event fusion based model. Extensive experiments show that the proposed approach outperforms state-of-the-art frame-based tracking methods by at least 10.4\% and 11.9\% in terms of representative success rate and precision rate, respectively. Besides, the effectiveness of each key component of our approach is evidenced by our thorough ablation study.}

\end{abstract}

\section{Introduction}
\bd{
Recently, convolutional neural networks (CNNs) based approaches show promising performance in object tracking tasks~\cite{bertinetto2016fully,bhat2020know, Chen_2020_CVPR, dai2019visual, danelljan2017eco,Gao_2020_CVPR, Guo_2020_CVPR,li2019gradnet,nam2016learning, zhang2021multi, zhang2018learning, zhang2019deeper}. 
These approaches mainly use conventional frame-based cameras as sensing devices since they can effectively measure absolute light intensity and provide a rich representation of a scene. However, conventional frame-based sensors have limited frame rates (\ie, $\le$ 120 FPS) and dynamic range (\ie, $\le$ 60 dB). Thus, they do not work robustly in degraded conditions. Figure~\ref{fig:main_page_figure} (a) and (b) show two examples of degraded conditions, high dynamic range, and fast-moving object, respectively. Under both conditions, we hardly see the moving objects. Thus, obtaining meaningful visual cues of the objects is challenging. By contrast, an event-based camera, a bio-inspired sensor, offers high temporal resolution (up to 1MHz), high dynamic range (up to 140 dB), and low energy consumption~\cite{brandli2014240}. Nevertheless, it cannot measure absolute light intensity and thus texture cues (as shown in Figure~\ref{fig:main_page_figure} (d)). Both sensors are, therefore, complementary. The unique complementarity triggers us to propose a multi-modal sensor fusion-based approach to improve the tracking performance in degraded conditions, which leverages the advantages of both the frame- and event-domain.
}
\def\wdenoising{0.9\linewidth}
\def\hdenoising{2.5in}
\begin{figure}[t]
	\setlength{\tabcolsep}{1.0pt}
	\centering
	\begin{tabular}{c}
		
		\includegraphics[width=\wdenoising]{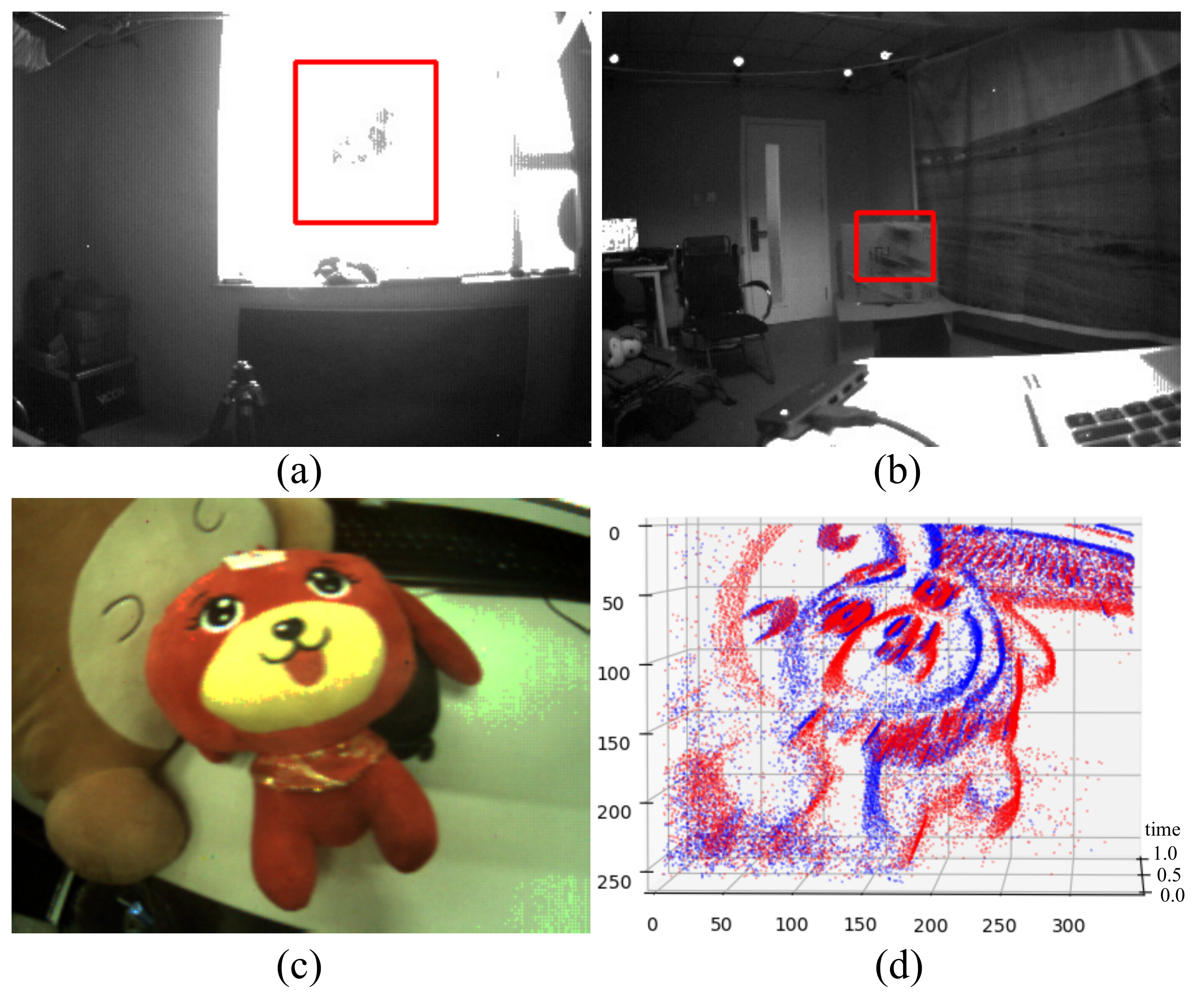} \\
		
	\end{tabular}
	\caption{\bd{\textbf{Limitations of conventional frame-based and bio-inspired event-based cameras.} (a) and (b) show the limitation of a frame-based camera under HDR and fast-moving object, respectively. (d) shows an event-based camera's asynchronous output of the scene shown in (c), sparse and no texture information.} }
	\label{fig:main_page_figure}
	\vspace{-0.35cm}
\end{figure}

\bd{
Yet, event-based cameras measure light intensity changes and output events asynchronously. It differs significantly from conventional frame-based cameras, which represent scenes with synchronous frames. Besides, CNNs-based approaches are not designed to digest asynchronous inputs. Therefore, combining asynchronous events and synchronous images remains challenging. To address the challenge, we propose a simple yet effective event aggregation approach to discretize the time domain of asynchronous events. Each of the discretized time slices can be accumulated to a conventional frame, thus can be easily processed by a CNNs-based model. Our experimental results show the proposed aggregation method outperforms other commonly used event accumulation approaches~\cite{chen2020end,lagorce2016hots,maqueda2018event,rebecq2017real,zhu2019unsupervised}. Another critical challenge, similar to other multi-modal fusion-based approaches~\cite{an2016online,camplani2015real, lan2018robust, li2018cross,  Mei_2021_CVPR_PDNet, song2013tracking}, is grasping meaningful cues from both domains effectively regardless the diversity of scenes. In doing so, we introduce a novel cross-domain feature integrator, which leverages self- and cross-domain attention schemes to fuse visual cues from both the event- and frame-domain effectively and adaptively. The effectiveness is enforced by a novel designed feature enhancement module, which enhances its own domain's feature based on both domains' attentions. Our approach's adaptivity is held by a specially designed weighting scheme to balance the contributions of the two domains. Based on the two domains' reliabilities, the weighting scheme adaptively regulates the two domains' contributions. We extensively validate our multi-modal fusion-based method and demonstrate that our model outperforms state-of-the-art frame-based methods by a significant margin, at least 10.4\% and 11.9\% in terms of representative success rate and precision rate, respectively.
}

\bd{
To exploit event-based visual cues in single object tracking and enable more future research on multi-modal learning with asynchronous events, we construct a large-scale single-object tracking dataset, FE108, which contains 108 sequences with a total length of 1.5 hours. FE108 provides ground truth annotations on both the frame- and event-domain. The annotation frequency is up to 40Hz and 240Hz for the frame and event domains, respectively. To the best of our knowledge, FE108 is the largest event-frame-based dataset for single object tracking, which also offers the highest annotation frequency in the event domain. 
}

\bd{To sum up, our contributions are as follows:}

\bd{$\bullet$ We introduce a novel cross-domain feature integrator, which can effectively and adaptively fuse the visual cues provided from both the frame and event domains.  }

\bd{$\bullet$ We construct a large-scale frame-event-based dataset for single object tracking. The dataset covers wide  challenging scenes and degraded conditions. }

\bd{$\bullet$ Our extensively experimental results show our approach outperforms other state-of-the-art methods by a significant margin. Our ablation study evidences the effectiveness of the novel designed attention-based schemes. }

\section{Related Work}

\noindent
\bd{ \tb{Single-Domain Object Tracking.} Recently, deep-learning-based methods have dominated the frame-based object tracking field. Most of the methods~\cite{bertinetto2016fully, dai2019visual, danelljan2017eco, li2019gradnet, nam2016learning,  zhang2018learning, zhang2019deeper} leverage conventional frame-based sensors. Only a few attempts have been made to track objects using event-based cameras. Piatkowska \textit{et al.}~\cite{piatkowska2012spatiotemporal} used an event-based camera for multiple persons tracking in the occurrence of high occlusions, which is enabled by the Gaussian Mixture Model based events clustering algorithm. Barranco \textit{et al.}~\cite{barranco2018real} proposed a real-time mean-shift clustering algorithm using events for multi-object tracking. Mitrokhin \textit{et al.}~\cite{mitrokhin2018event} proposed a novel events representation, time-image, to utilize temporal information of the event stream. With it, they achieve an event-only feature-less motion compensation pipeline.
Chen \textit{et al.}~\cite{chen2020end} pushed the event representation further and proposed a synchronous Time-Surface with Linear Time Decay representation to effectively encode the Spatio-temporal information. Although these approaches reported promising performance in object tracking tasks, they did not consider leveraging frame domains.  By contrast, our approach focuses on leveraging complementarity between frame and event domains. }

\noindent
\bd{\tb{Multi-Domain Object Tracking.} Multi-modal-based tracking approaches have been getting more attention. Most of the works leverage RGB-D (RGB + Depth)~\cite{an2016online,camplani2015real, kart2018make, song2013tracking,   xiao2017robust} and RGB-T (RGB + Thermal)~\cite{lan2018robust, li2018cross, li2019multi, Wang_2020_CVPR, zhang2019multi, zhu2019dense} as multi-modal inputs to improve tracking performance. Depth is an important cue to help solve the occlusion problem in tracking. When a target object is hidden partially by another object with a similar appearance, the difference in their depth levels will be distinctive and help detect the occlusion. Illumination variations and shadows do not influence images from the thermal infrared sensors. They thus can be combined with RGB to improve performance in degraded conditions (\eg, rain and smog). Unlike these multi-domain approaches, fusing frame and event domains brings a unique challenge caused by the asynchronous outputs of event-based cameras. Our approach aims to solve the problem, which effectively leverage events for improving robustness, especially under degraded conditions.}

\section{Methodology}

\subsection{Background: Event-based Camera}
\bd{An event-based camera is a bio-inspired sensor. It asynchronously measures light intensity changes in scene-illumination at a pixel level. Hence, it provides a very high measurement rate, up to 1MHz~\cite{brandli2014240}. Since light intensity changes are measured in log-scale, an event-based camera also offers a very high dynamic range, 140 dB vs. 60 dB of a conventional camera~\cite{DBLP:journals/corr/abs-1904-08405}. When the change of pixel intensity in the log-scale is greater than a threshold, an event is triggered.  The polarity of an event reflects the direction of the changes. Mathematically, a set of events can be defined as:}
\bd{\begin{equation} 
\label{eq:event_set}
\mathcal{E} = \{e_k\}_{k=1}^N = \{[x_k, y_k, t_k, p_k]\}_{k=1}^N,
\end{equation}
where $e_k$ is the $k$-th event; $(x_k, y_k)$ is the pixel location of event $e_k$; $t_k$ is the timestamp; $p_k \in \{-1, 1\}$ is the polarity of an event. In a stable lighting condition, events are triggered by moving edges (\eg, object contour and texture boundaries), making an event-based camera a natural edge extractor.  }


\subsection{Event Aggregation}
\label{subsec:EA}
\bd{Since the asynchronous event format differs significantly from the frames generated by conventional frame-based cameras, vision algorithms designed for frame-based cameras cannot be directly applied. To deal with it, events are typically aggregated into a frame or grid-based representation first~\cite{ gehrig2019end,lagorce2016hots, maqueda2018event,messikommer2020event,rebecq2017real,wang2021event,zhu2019unsupervised}. }

\bd{We propose a simple yet effective pre-processing method to map events into a grid-based presentation. Specifically, inspired by \cite{zhu2019unsupervised}, we first aggregate the events captured between two adjacent frames into an $n$-bin voxel grid to discretize the time dimension. Then, each 3D discretized slice is accumulated to a 2D frame, where a pixel of the frame records the polarity of the event with the latest timestamp at the pixel's location inside the current slice. Finally, the $n$ generated frames are scaled by $255$ for further processing. Given a set of events, $\mathcal{E}^i = \{e_k^i\}_{k=1}^{N_i}$, with the timestamps in the time range of $i$-th bin, the pixel located at $(x, y)$ on the $i$-th aggregated frame can be defined as follows:}
\begin{multline}
\label{eq:ep}
g(x, y, i) =  \lfloor\frac{p_k^i\times\delta(t(x, y, i)_{max} - t_k^i) + 1}{2} \times 255\rfloor\\ 
t(x, y, i)_{max} = max(t_k^i\times \delta(x-x_k^i, y-y_k^i)) \\ 
\forall t_k^i \in [T_j + (i-1) B, T_j + i B],
\end{multline}
\bd{where $T_j$ is the timestamp of the $j$-th frame in the frame domain; $\delta$ is the Dirac delta function; $B$ is the bin size in the time domain, which is defined as: $B = (T_{j+1} - T_j) / n$. The proposed  method leverages the latest timestamp to capture the latest motion cues inside each time slice. Our experimental results show that our event processing method outperforms other commonly used approaches (see Table~\ref{tab:ablation}).}

\subsection{Network Architecture}

\bd{The overall architecture of the proposed approach is illustrated in Figure~\ref{fig:overview}, which has two branches: reference branch (top) and test branch (bottom). The reference and test branches share weights in a siamese style. 
The core of our approach is the Cross-Domain Feature Integrator (CDFI), designed to leverage both domains' advantages. Specifically, the frame domain provides rich texture information, whereas the event domain is robust to challenging scenes and provides edge information. As shown in Figure~\ref{fig:overview}, the inputs of CDFI are a frame and events captured between the frame and its previous one. We preprocess the events based on Eq.~\ref{eq:ep}. The outputs of CDFI are one low-level (\ie, $K_l$) and one high-level (\ie, $K_h$) fused features. The classifier uses the extracted low-level fused features from both reference and test branches to estimate a confidence map. Finally, the bbox-regressor reports IoU between the ground truth bounding box and estimated bounding box to help locate a target on the test frame. }

\def\wdenoising{1.0\linewidth}
\def\hdenoising{2.0in}
\begin{figure}[t]
	\setlength{\tabcolsep}{1.0pt}
	\centering
	\begin{tabular}{c}
		
		\includegraphics[width=\wdenoising]{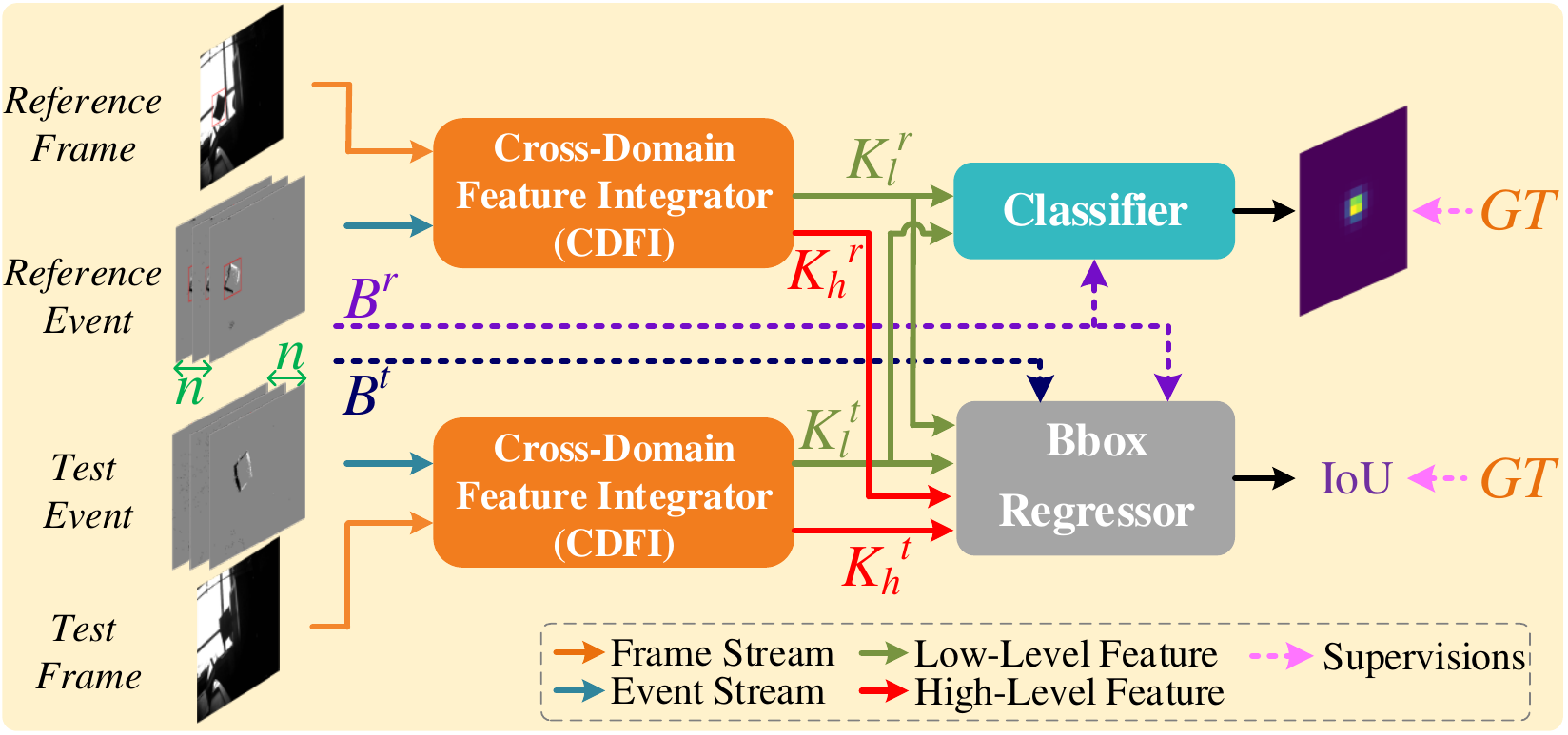} \\
		
	\end{tabular}
	\caption{\bd{Overview of the proposed architecture.}}
	\label{fig:overview}
	\vspace{-0.35cm}
\end{figure}

\def\wdenoising{1.0\linewidth}
\def\hdenoising{3.75in}
\begin{figure*}[t]
	\setlength{\tabcolsep}{1.0pt}
	\centering
	\begin{tabular}{c}
		
		\includegraphics[width=\wdenoising, height=\hdenoising]{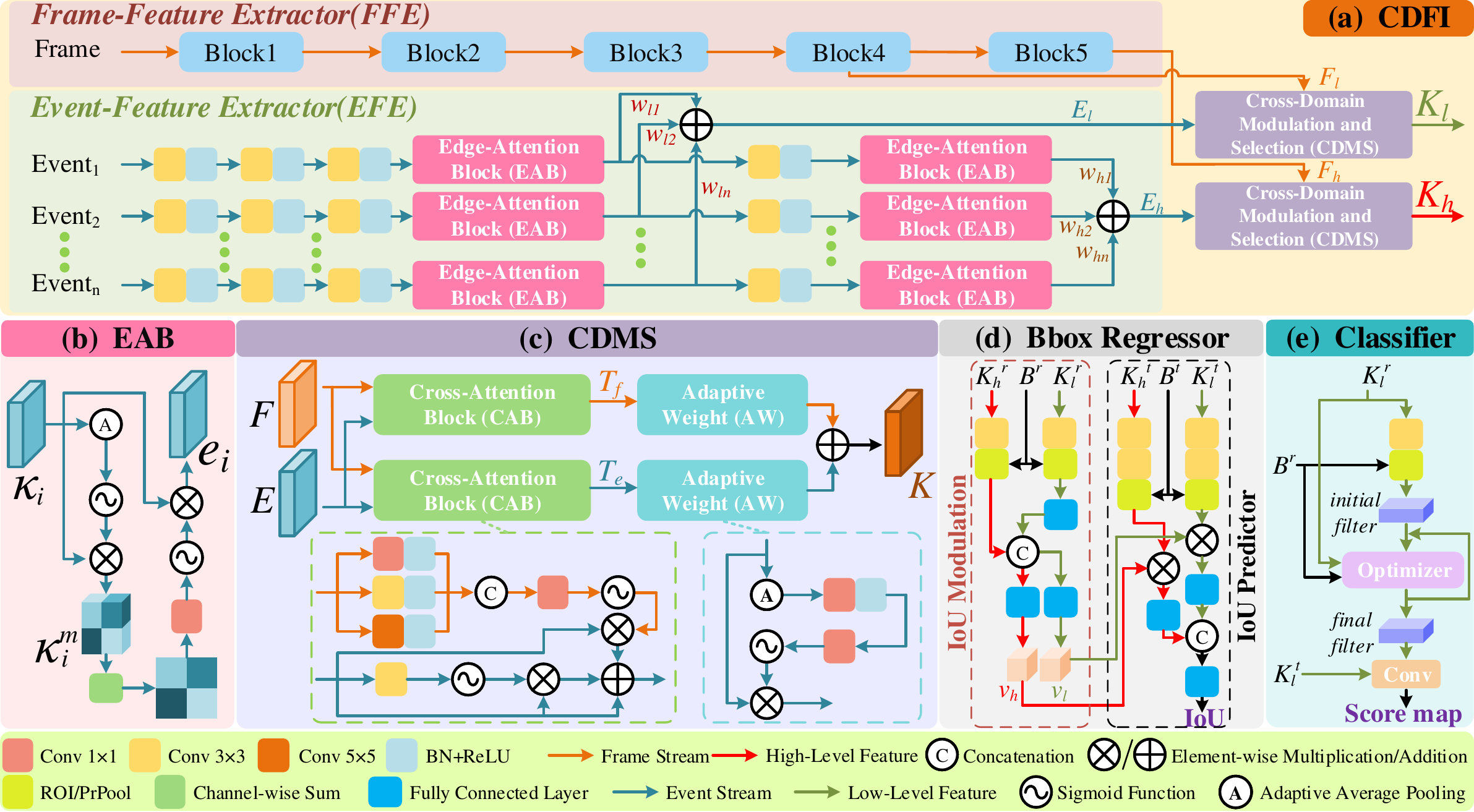} \\
		
	\end{tabular}
	\caption{\bd{Detailed architectures of the proposed components. (a) Overview of Cross-Domain Feature Integrator (CDFI), (b) Edge-Attention Block (EAB), (c) Cross-Domain Modulation and Selection block (CDMS), (d) Bbox Regressor, and (e) Classifier.}}
	\label{fig:pipeline}
	\vspace{-0.35cm}
\end{figure*}

\subsubsection{Cross-Domain Feature Integrator}
\bd{The overall structure of the proposed CDFI is shown in Figure~\ref{fig:pipeline} (a). It has three components, namely: Frame-Feature Extractor (FFE), Event-Feature Extractor (EFE), and Cross-Domain Modulation and Selection Block (CDMS). }

\noindent \bd{\tb{FFE} is for extracting features from the frame domain. We adopt ResNet18~\cite{he2016deep} as our frame feature extractor. The 4th and 5th blocks' features are used as the low-level and high-level frame features (\ie, $F_l$ and  $F_h$), respectively. }

\noindent \bd{\tb{EFE} generates features to represent the encoded information in the event domain. Similar to FFE, EFE extracts low-/high-level features from the event domain (\ie, $E_l$ and $E_h$). Since each aggregated event frame conveys different temporal information, each of them is processed by a dedicated sub-branch. Like other feature extractors, each sub-branch of EFE leverages stacked convolutional layers to increase receptive field at higher levels. We also introduce a self-attention scheme to each sub-branch to focus on more critical features. It is achieved by a specially designed Edge Attention Block (EAB), illustrated in Figure~\ref{fig:pipeline} (b). As shown in Figure~\ref{fig:pipeline} (a), two EABs are added behind the third and fourth convolutional layers. Then, the low-level (\ie, $e_i^l$) and high-level (\ie, $e_i^h$) features on the $i$th sub-branch are generated by the first and second EABs, respectively. Finally, all generated $e_i^l$ and $e_i^h$ are fused in a weighted sum manner to obtain the $E_l$ and $E_h$. Mathematically, EFE is defined as (here we ignore $l$ and $h$ to bring a general form):
\begin{align}
\label{equ:weight}
E &= w_{1}e_{1} \oplus ... \oplus w_{n}e_{n},\\
e_i &= \sigma(\psi_{1\times1}(\mathcal{C}(\kappa_{i}^m))) \otimes \kappa_i,\\ 
\kappa_{i}^m &= \sigma(\mathcal{A}(\kappa_i)) \otimes \kappa_i,
\end{align}
where $w_i$ is a learned weight; $\psi_{1\times 1}$ means a $1\times1$ convolution layer; $\sigma$ is the Sigmoid function; $\kappa_i$, $e_i$, $\mathcal{C}$, and $\mathcal{A}$ are the input, output features of the EAB on the $i$th sub-branch, channel-wise addition, and adaptive average pooling, respectively; $\oplus$/$\otimes$ indicates element-wise summation/multiplication; 
}

\noindent
\bd{\tb{CDMS} is designed to fuse the extracted frame and event features, shown in Figure~\ref{fig:pipeline} (c). The key to the proposed CDMS is a cross-domain attention scheme designed based on the following observations: (i) Rich textural and semantic cues can easily be captured by a conventional frame-based sensor, whereas an event-based camera can easily capture edge information. (ii) The cues provided by a conventional frame-based sensor become less effective in challenging scenarios. By contrast, an event-based camera does not suffer from these scenarios. (iii) In the case of multiple moving objects crossing each other, it is hard to separate them trivially based on edges. However, the problem can be addressed well with texture information.}

\bd{To address the first observation, we design a Cross-Attention  Block (CAB) to fuse features of the two domains based on cross-domain attentions. Specifically, given two features from two different domains, $D_1$ and $D_2$, we define the following cross-domain attention scheme to generate an enhanced feature for $D_1$:}
\begin{align}
\label{eq:CDMS_CAB}
T_{D_1} &= T_{D_1}^{1\rightarrow{1}} \oplus  T_{D_1}^{2\rightarrow 1} \oplus D_1  \\
T_{D_1}^{1\rightarrow{1}} &= \sigma(\psi_{3\times3}(D_1)) \otimes D_1, \label{eq:CDMS_SA}\\ 
\begin{split}
T_{D_1}^{2\rightarrow 1} &= \sigma(\psi_{1\times1}[\xi(\psi_{1\times1}(D_2)),\\
&\quad \ \ \xi(\psi_{3\times3}(D_2)) , \xi(\psi_{5\times5}(D_2))])\otimes D_1, \label{eq:CDMS_CA}
\end{split}
\end{align}
\bd{where    $[\cdot]$ indicates channel-wise concatenation; $\xi$ is the Batch Normalization
(BN) followed by a ReLU activation function; 
$T_{D_1}^{1\rightarrow 1}$ indicates a self-attention based on $D_1$. $T_{D_1}^{2\rightarrow 1}$ is a cross-domain attention scheme based on $D_2$ to enhance the feature of $D_1$.
When $D_1$ and $D_2$ represent the event- and frame-domain, the enhanced feature of the event-domain, $T_e$, is obtained. Inversely, the enhanced feature of the frame-domain, $T_f$, can be generated. }

\bd{To address the second and third observations, we propose an adaptive weighted balance scheme to balance the contribution of the frame- and event- domains:
\begin{align}
\label{eq:CDMS_aw}
K &= W_{f}T_{f} \oplus W_{e}T_{e},\\
W_{D_i} &= \sigma(\psi_{1\times1}(\xi(\psi_{1\times1}(\mathcal{A}(T_{D_i}))))).
\end{align}
}

\vspace{-0.6cm}
\subsubsection{Bounding Box (BBox) Regressor and Classifier}
\bd{
For the BBox regressor and classifier, we adopt the target estimation network of ATOM\cite{danelljan2019atom} and the classifier of DiMP~\cite{bhat2019learning}, respectively. The architecture of BBox regressor is shown in Figure~\ref{fig:pipeline} (d). The IoU modulation maps $K_{l}^{r}$ and $K_{h}^{r}$ to different level modulation vectors $v_{l}$ and $v_{h}$, respectively. Mathematically, the mapping is achieved as follows:
\begin{equation} 
\begin{aligned}
& \qquad v_{l} = \mathcal{F}(q), \qquad  v_{h} = \mathcal{F}(q),  \\
q = [ &\mathcal{F}(\mathcal{P}(\psi_{3\times3}(K_{l}^{r}), B^r)), \mathcal{P}(\psi_{3\times3}(K_{h}^{r}), B^r)] 
\end{aligned}
\end{equation}
where $\mathcal{F}$ is fully connected layer; $\mathcal{P}$ denotes PrPool~\cite{jiang2018acquisition}; $B^r$ is the target bounding box from reference branch. 
The IoU predictor predicts IoU based on the following equation:
\begin{equation} 
\begin{aligned}
IoU =\mathcal{F}([&\mathcal{F}(\mathcal{P}(\psi_{3\times3}(\psi_{3\times3}(K_{l}^{t})), B^t) \otimes v_{l}), \\
&\mathcal{F}(\mathcal{P}(\psi_{3\times3}(\psi_{3\times3}(K_{h}^{t})), B^t) \otimes v_{h} )])
\end{aligned}
\end{equation}
}

\bd{
For the classifier, following~\cite{bhat2019learning}, we use it to predict a target confidence score. As shown in Figure~\ref{fig:pipeline} (e),  the classifier first maps $K_{l}^{r}$ and $B^r$ to an initial filter, which is then optimized by the optimizer. The optimizer uses the steepest descent methodology to obtain the final filter. The final filter is used as the convolutional layer's filter weight and applied to $K_{l}^{t}$ to robustly discriminate between the target object and background distractors.}

\subsection{Loss Function}
\bd{
We adopt the loss function of~\cite{bhat2019learning}, which is defined as:
\begin{align}
\label{eq:Loss}
L_{\mathrm{tot}} &=\beta L_{\mathrm{cls}}+L_{\mathrm{b}}, \\
L_{c l s} &= \frac{1}{N} \sum_{i=1}^{N}(\ell\left(s_i, z_{c}\right))^{2},\\ 
&\ell(s_i, z_{c}) =\left\{\begin{array}{ll}
s_i-z_{c}, & z_{c}>0.05 \\
\max (0, s_i), & z_{c} \leq 0.05,
\end{array}\right. \label{eq:loss_hinge}\\
L_b&=\frac{1}{N} \sum_{i=1}^{N}(IoU_{i}-IoU_{gt})^{2},
\end{align}
where $s_i$ is the $i$-th classification score predicted by the classifier, and $z_c$ is obtained by setting to a Gaussian function centered as the target $c$. The loss function has two components: classification loss $L_{\mathrm{cls}}$, and bounding box regressor loss $L_b$. The $L_{\mathrm{cls}}$ estimates Mean Squared Error (MSE) between $s_i$ and $z_c$. The idea behind Eq.~\ref{eq:loss_hinge} is to alleviate the impact of unbalanced negative samples (\ie, background). A hinge function is applied to clip the scores at zero in the background region so that the model can equally focus on both positive and negative samples. The $L_b$ estimates MSE between the predicted IoU overlap $IoU_{i}$ obtained from test branch and the ground truth $IoU_{gt}$. 
}

\section{Dataset}
\bd{
Currently, Hu \textit{et al.}~\cite{hu2016dvs} collected a dataset by placing an event-based camera in front of a monitor and recorded large-scale annotated RGB/grayscale videos (\textit{e.g.}, VOT2015~\cite{kristan2015visual}). However, the dataset based on RGB tracking benchmarks cannot faithfully represent events captured in real scenes since the events between adjacent frames are missing. 
Mitrokhin \textit{et al.}~\cite{mitrokhin2018event,mitrokhin2019ev} collected two event-based tracking datasets: EED~\cite{mitrokhin2018event} and EV-IMO~\cite{mitrokhin2019ev}. As shown in Table~\ref{tab:dataset}, the EED only has 179 frames (7.8 seconds) with two types of objects. EV-IMO offers a better package with motion masks and high-frequency events annotations, up to 200Hz. But, similar to EED, limited object types block it to be used practically. To enable further research on multi-modal learning with events, we collect a large-scale dataset termed FE108, which has 108 sequences with a total length of 1.5 hours. The dataset contains 21 different types of objects and covers four challenging scenarios. The annotation frequency is up to 20/40 Hz for the frame domain (20 out of 108 sequences are 20Hz) and 240 Hz for the event domain. 
}

\setlength{\tabcolsep}{0.55pt}
\begin{table}[t]
	\centering
	
	    \scalebox{0.9}{
	\begin{tabular}{lcccccc}
		\hline
		\hline
		 & Classes  &    Frames     &   Events    &    Time    &   Frame(Hz)    &  Event(Hz)   \\ 
		 \hline
		 EED~\cite{mitrokhin2018event}      &      2      &       179        &      3.4M       &       7.8s        &       23       &      23    \\
		EV-IMO~\cite{mitrokhin2019ev} &      3      &      76,800      &       --        &     32.0m      &       40      &     200     \\
		Ours            & \textbf{21} & \textbf{208,672} & \textbf{ 5.6G } & \textbf{96.9m} & \textbf{20/40} & \textbf{240} \\ \hline
		\hline
	\end{tabular}}
	\caption{\bd{\textbf{Analysis of existing event-based datasets.} Our FE108 offers the best in terms of all listed metrics.}}
    \label{tab:dataset}
	\vspace{-0.35cm}
\end{table}

	
	
		 
		

\subsection{Dataset Collection and Annotation}
\bd{
The FE108 dataset is captured by a DAVIS346 event-based camera~\cite{brandli2014240}, which equips a 346$\times$260 pixels dynamic vision sensor (DVS) and an active pixel sensor (APS). It can simultaneously provide events and aligned grayscale images of a scene. The ground truth bounding boxes of a moving target are provided by the Vicon motion capture system~\cite{vicon}, which captures motion with a high sampling rate (up to 330Hz) and sub-millimeter precision. During the capturing process, we fix APS's frame rate to 20/40 FPS and Vicon's sampling rate to 240Hz, which are also the annotation frequency of the captured APS frame and accumulated events (\ie, accumulated every $1/240$ second), respectively.
}

\subsection{Dataset Facts}
\label{subsec:dataset_facts}
\bd{
We introduce critical aspects of the constructed FE108. More details about the FE108 are described in the \textit{supplementary material}.
}

\noindent
\bd{
\tb{Categorical Analysis.} The FE108 dataset can be categorized differently from different perspectives. The first perspective is the number of object classes. There are 21 different object classes, which can be divided into three categories: animals, vehicles, and daily goods (\textit{e.g.,} bottle, box). Second, as shown in Figure~\ref{fig:statistics} (a), the FE108 contains four types of challenging scenes: low-light (LL), high dynamic range (HDR), fast motion with and without motion blur on APS frame (FWB and FNB). Third, based on the camera movement and number of objects, FE108 has four types of scenes: static shots with a single object or multiple objects; dynamic shots with a single object or multiple objects.
}

\noindent
\bd{
\tb{Annotated Bounding Box Statistics.} In Figure~\ref{fig:statistics} (b), we plot out the distribution of all annotated bounding box locations, which shows most annotations are close to frames' centers. In Figure~\ref{fig:statistics} (c), we also show the distribution of the bounding box aspect ratios (H/W) . 
}

\noindent
\bd{
\tb{Event Rate.} The FE108 dataset is collected in a constant lighting condition. It means all events are triggered by motions (\eg, moving objects, camera motion). Therefore, the distribution of the event rate can represent the motion distribution of FE108. As shown in Figure~\ref{fig:statistics} (d), the distribution of the event rate is diverse. It indicates the captured 108 scenes offer wide motion diversity. }


\def\wdenoising{0.49\linewidth}
\def\hdenoising{1.1in}
\begin{figure}[t]
	\setlength{\tabcolsep}{1.2pt}
	\centering
	
	\begin{tabular}{cc}
	
		\includegraphics[width=\wdenoising, height=\hdenoising]{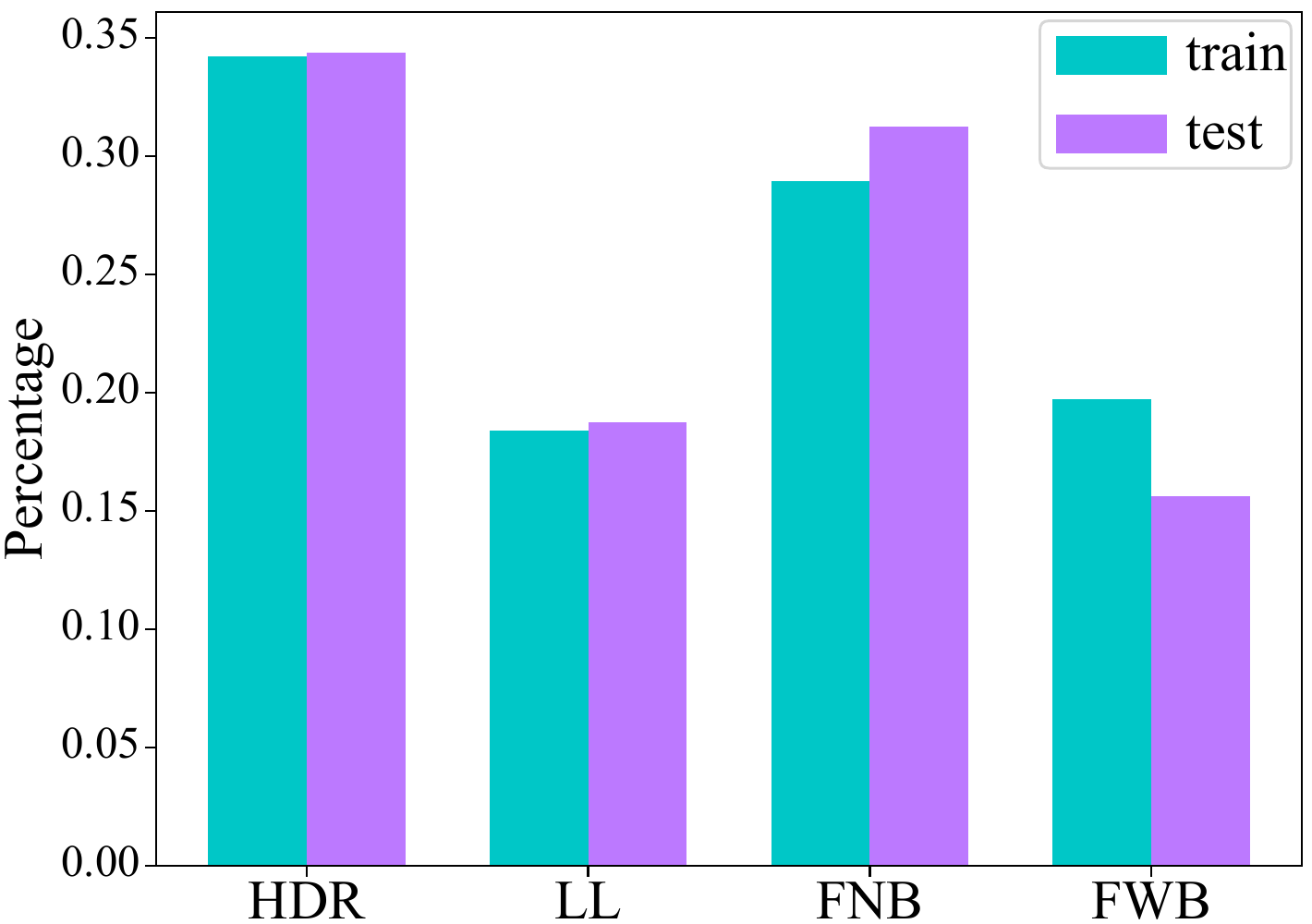} & 
		\includegraphics[width=\wdenoising, height=\hdenoising]{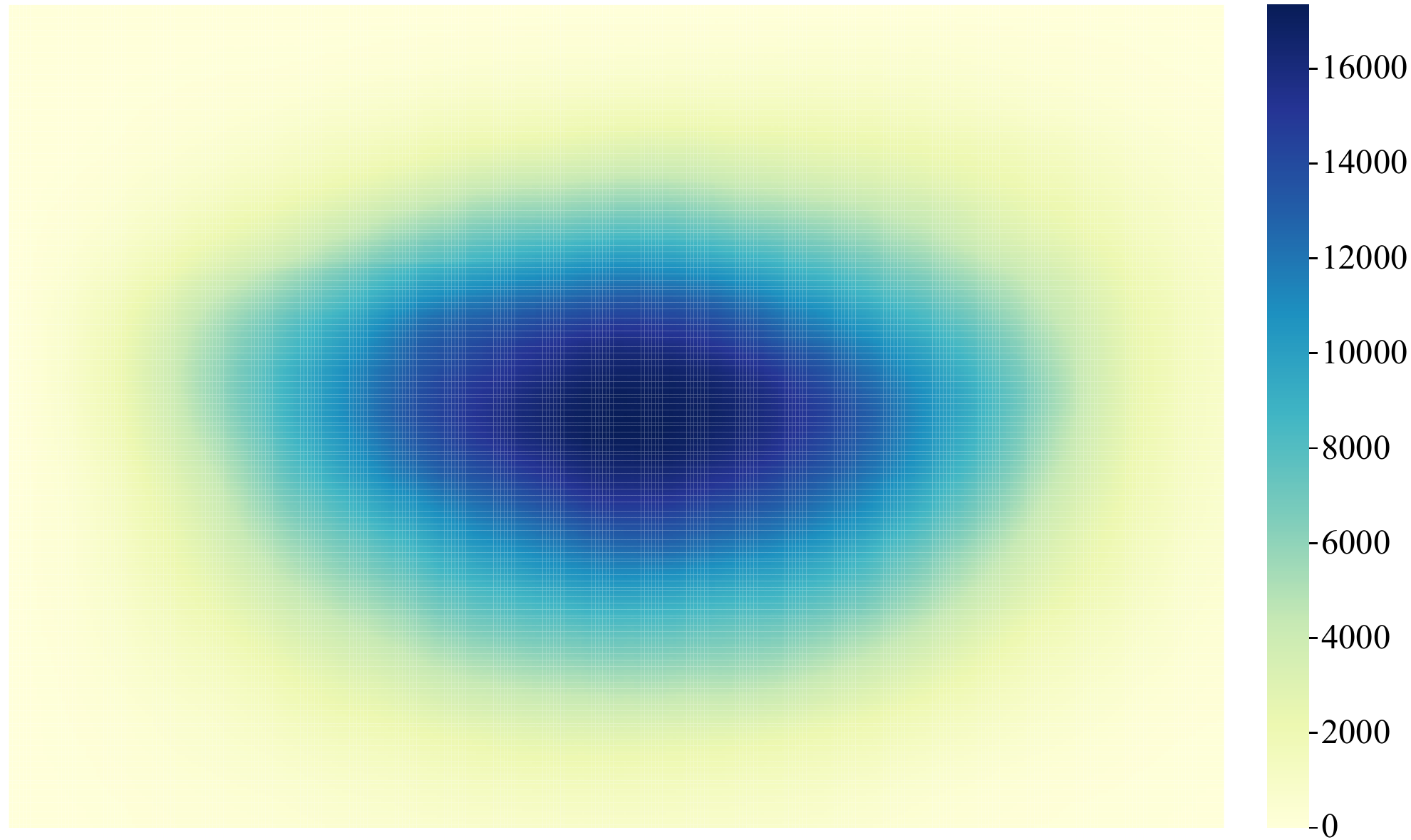}  \\
		{\small (a) Attributes distribution} & {\small (b) Bounding box distribution}  \\
		\includegraphics[width=\wdenoising, height=\hdenoising]{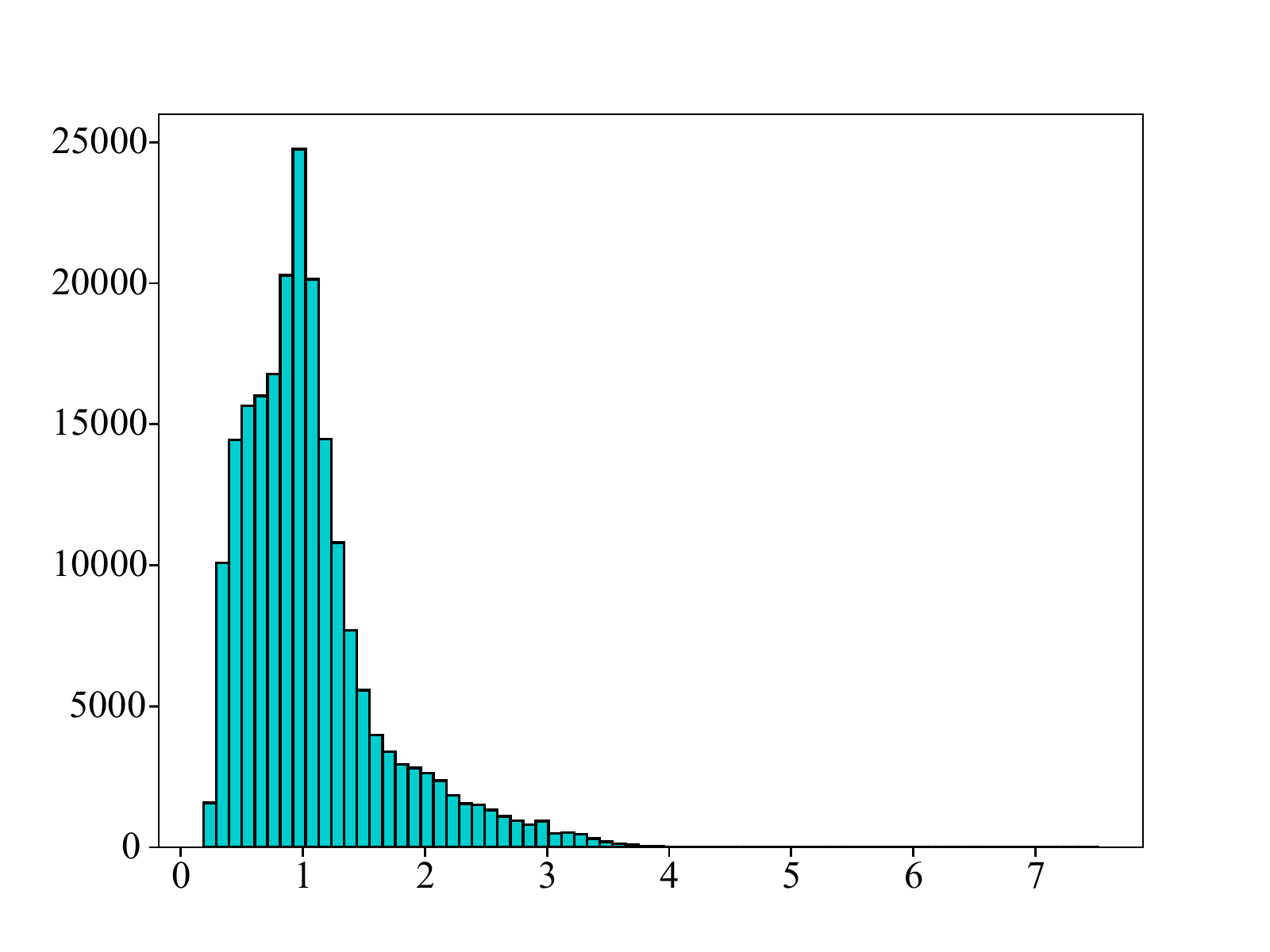}  &
		 \includegraphics[width=\wdenoising, height=\hdenoising]{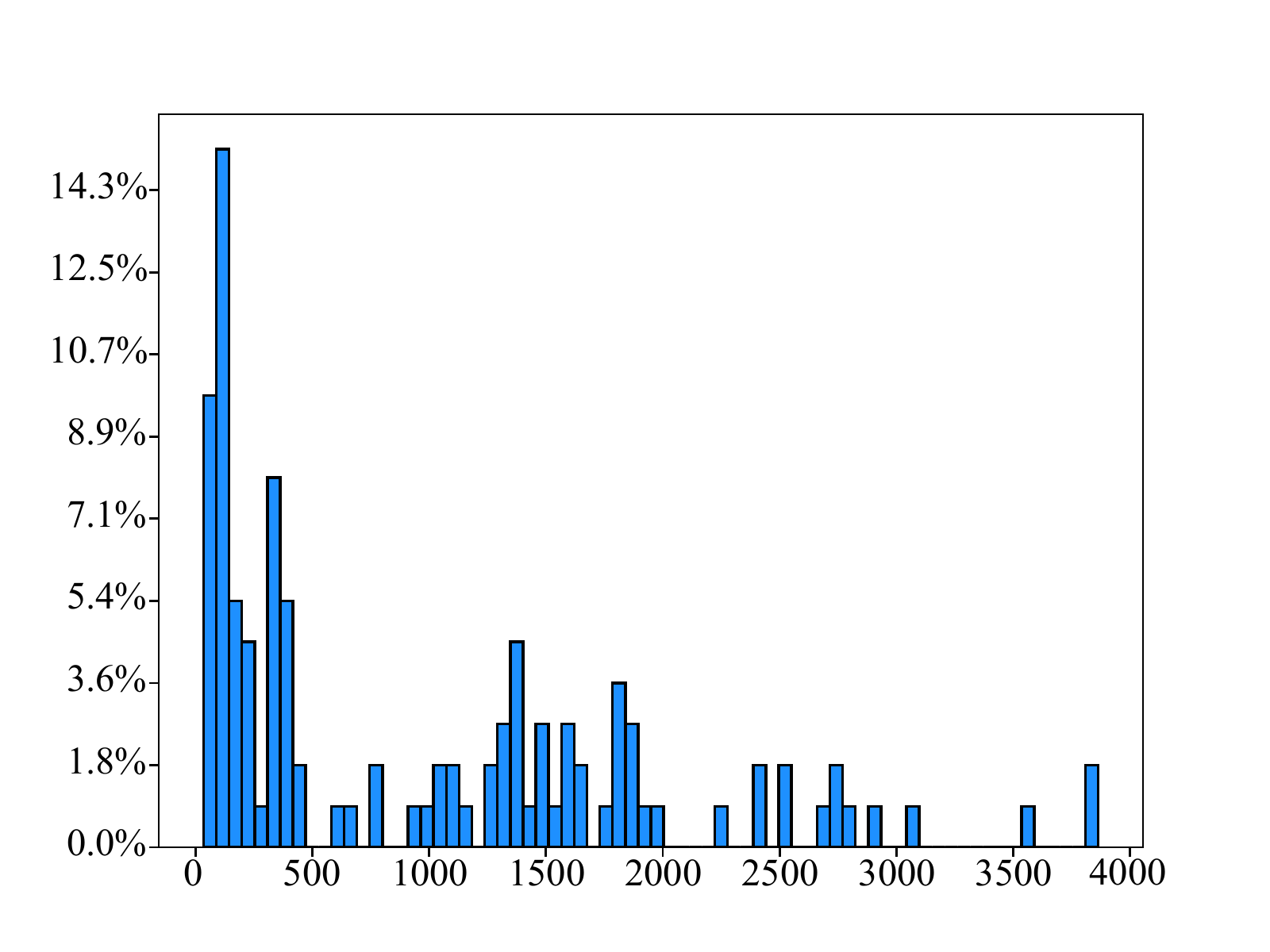} \\
		{\small (c) Histogram of aspect ratios} & {\small (d)  Avg event rate (Ev/ms)}  \\
		
	\end{tabular}
	\caption{\bd{Statistics of FE108 dataset in terms of (a) attributes, (b) bounding box center position, (c) aspect ratios, and (d) event rate.}}
	\label{fig:statistics}
	\vspace{-0.35cm}
\end{figure}

\section{Experiments}

\bd{We implement the proposed network in PyTorch~\cite{paszke2019pytorch}. In the training phase, random initialization is used for all components except the FFE (which is a ResNet18 pre-trained on ImageNet). The initial learning rate for the classifier, the bbox regressor, and the CDFI are set to 1e-3, 1e-3, and 1e-4, respectively. The learning rate is adjusted by a decay scheduler, which is scaled by 0.2 for every 15 epochs. We use Adam optimizer to train the network for 50 epochs. The batch size is set to 26. It takes about 20 hours on a 20-core i9-10900K 3.7 GHz CPU, 64 GB RAM, and an NVIDIA RTX3090 GPU.}

\setlength{\tabcolsep}{1.2pt}
\begin{table*}[t]
	\centering
	\small
	
		    \scalebox{0.9}{
	\begin{tabular}{l|cccc|cccc|cccc|cccc|cccc}
		\hline
		\hline
		\multirow{2}{*}{methods} & \multicolumn{4}{c|}{HDR} & \multicolumn{4}{c|}{LL} & \multicolumn{4}{c|}{FWB}   & \multicolumn{4}{c|}{FNB}     & \multicolumn{4}{c}{ALL} \\
		\cline{2-21} 
		& RSR & OP$_{0.50}$ &OP$_{0.75}$ & RPR     & RSR &OP$_{0.50}$ &OP$_{0.75}$    & RPR       & RSR &OP$_{0.50}$ &OP$_{0.75}$  & RPR   & RSR &OP$_{0.50}$ &OP$_{0.75}$    & RPR   & RSR &OP$_{0.50}$ &OP$_{0.75}$ & RPR  \\ 
		\hline
		SiamRPN~\cite{li2018high} &15.3   &16.9   &6.1    &21.6 &10.1   & 8.3  & 1.4   &14.5 &26.2  & 32.1  &6.1    &44.1 &33.2  &42.9   &11.5    &51.9 &21.8  & 26.1  &7.0    &33.5 \\ 
		ATOM~\cite{danelljan2019atom}  &36.6   &41.8   &14.4    &56.0 &28.6  &29.1   &5.8    &45.0 &66.8  &89.6   &32.6    &96.7 &57.1  &71.0   &28.0    &88.6 &46.5  &56.4   &20.1    &71.3 \\
		DiMP~\cite{bhat2019learning}   &41.8   &50.0  &17.9    &62.7 &45.6  &52.8  &11.2    &69.5 &69.4  & \textbf{94.7}  & 37.1   &99.7 &60.5  &75.6   &29.3    &93.2 &52.6  &65.4   &23.4    &79.1  \\
		SiamFC++~\cite{xu2020siamfc++}  &15.3   &15.0   &1.3    &25.2 &13.4  & 8.7  &0.8    &15.3 &28.6  & 36.3  &6.0    &48.2 &36.8  & 42.7  &7.4    &63.1 &23.8   &26.0   &3.9   &39.1 \\
		SiamBAN~\cite{chen2020siamese}  &16.3   &16.4   &3.9    &26.6 &15.5  & 14.8  &2.3    &26.5 &25.2  &26.3   &5.8    &46.7 &32.0  & 39.6  & 9.1   &51.4 &22.5  &25.0   &5.6    &37.4  \\
		KYS~\cite{bhat2020know}  &15.7   & 14.5  &5.2    &23.0 &12.0  & 8.0  &1.1    &18.0 &47.0  &63.9   &14.8    &73.3 &36.9  &44.5   &15.2    &57.9 &26.6  & 30.6  &9.2    &41.0  \\
		CLNet~\cite{dongclnet}  &30.0   &33.5   &9.6    &48.3 &13.7  & 6.0  &0.9    &23.6 &52.9  & 71.2  &23.3    &80.3 &40.8  & 46.3  &14.2    &67.7 &34.4  & 39.1  &11.8    &55.5  \\
		PrDiMP~\cite{danelljan2020probabilistic}   &44.3   &52.8   &19.6    &66.3 &44.6  &48.2   &8.9    &69.5 &67.0  & 89.9  &33.6    &99.7 &60.6  & 75.8  &\textcolor{blue}{29.7}    &93.3 &53.0   &65.0   &23.3   &80.5 \\
		
		\hline
		ATOM~\cite{danelljan2019atom} + Event &49.0  &59.2   &21.0    &68.8 &50.8   & 67.8  & 27.7   &72.6 &68.5  &90.4  &42.0    &97.2 &57.4  &71.1   &28.3    &90.2 &55.5  & 70.0  &27.4    &81.8 \\ 
		DiMP~\cite{bhat2019learning} + Event &50.1   &60.2   &23.7    &74.8 &57.0   &70.4  &28.2   &82.8 &\textcolor{blue}{70.1}  &\textcolor{blue}{94.2}  &44.2    &\textcolor{blue}{99.9} &60.8  &75.9   &29.1    &\textcolor{blue}{93.6} &57.1  & 71.2  &28.6    &85.1 \\ 
		PrDiMP~\cite{danelljan2020probabilistic} + Event &\textcolor{blue}{53.1}   &\textcolor{blue}{65.3}   &\textcolor{blue}{24.9}    &\textcolor{blue}{79.1} &\textcolor{blue}{60.3}   & \textcolor{blue}{79.6}  & \textcolor{blue}{29.8}   &\textcolor{blue}{90.5} &70.0  & 93.8  &\textcolor{blue}{44.8} &99.8   &\textcolor{blue}{61.8} &\textcolor{blue}{76.3}  &29.4   &\textcolor{blue}{93.6}   &\textcolor{blue}{59.0}  & \textcolor{blue}{74.4}  &\textcolor{blue}{29.8}    &\textcolor{blue}{87.7} \\ 
		
		\hline
		Ours  &\textbf{59.9}   & \textbf{74.4}  &\textbf{33.0} &\textbf{86.0} &\textbf{65.6}  & \textbf{86.0}  & \textbf{30.8}  &\textbf{95.7} &\textbf{71.2}  & \textbf{94.7}  &\textbf{45.9} &\textbf{100.0} &\textbf{62.8}  & \textbf{80.5}  & \textbf{32.0}  &\textbf{94.5} &\textbf{63.4}   & \textbf{81.3} &\textbf{34.4} &\textbf{92.4}  \\
		\hline
		\hline
	\end{tabular}}
	\caption{\bd{State-of-the-art comparison on FE108 in terms of representative success rate (RSR), representative precision rate (RPR), and overlap precision (OP).}}
	\label{tab:attribute}
	\vspace{-0.35cm}
\end{table*}

\subsection{Comparison with State-of-the-art Trackers}
\bd{To validate the effectiveness of our method, we compare the proposed approach with the following eight state-of-the-art frame-based trackers: SiamRPN~\cite{li2018high}, ATOM~\cite{danelljan2019atom}, DiMP~\cite{bhat2019learning}, SiamFC++~\cite{xu2020siamfc++}, SiamBAN~\cite{chen2020siamese}, KYS~\cite{bhat2020know}, CLNet~\cite{dongclnet}, and PrDiMP~\cite{danelljan2020probabilistic}. To show the quantitative performance of each tracker, we utilize three widely used metrics: success rate (SR), precision rate (PR), and overlap
precision ($\text{OP}_\text{T}$). These metrics represent the percentage of three particular types of frames. SR cares the frame of that
overlap between ground truth and predicted bounding box is larger than a threshold; PR focuses on the frame of that the center distance
between ground truth and predicted bounding box within a given threshold; $\text{OP}_\text{T}$ represents SR with $T$ as the threshold. For SR, we employ the area under curve (AUC) of an SR plot as representative SR (RSR). For PR, we use the PR score associated with a 20-pixel threshold as representative PR (RPR). }

\def\wdenoising{1.0\linewidth}
\def\hdenoising{1.2in}
\begin{figure}[tbp]
	\setlength{\tabcolsep}{1.0pt}
	\centering
	\small
	\begin{tabular}{c}

		\includegraphics[width=\wdenoising]{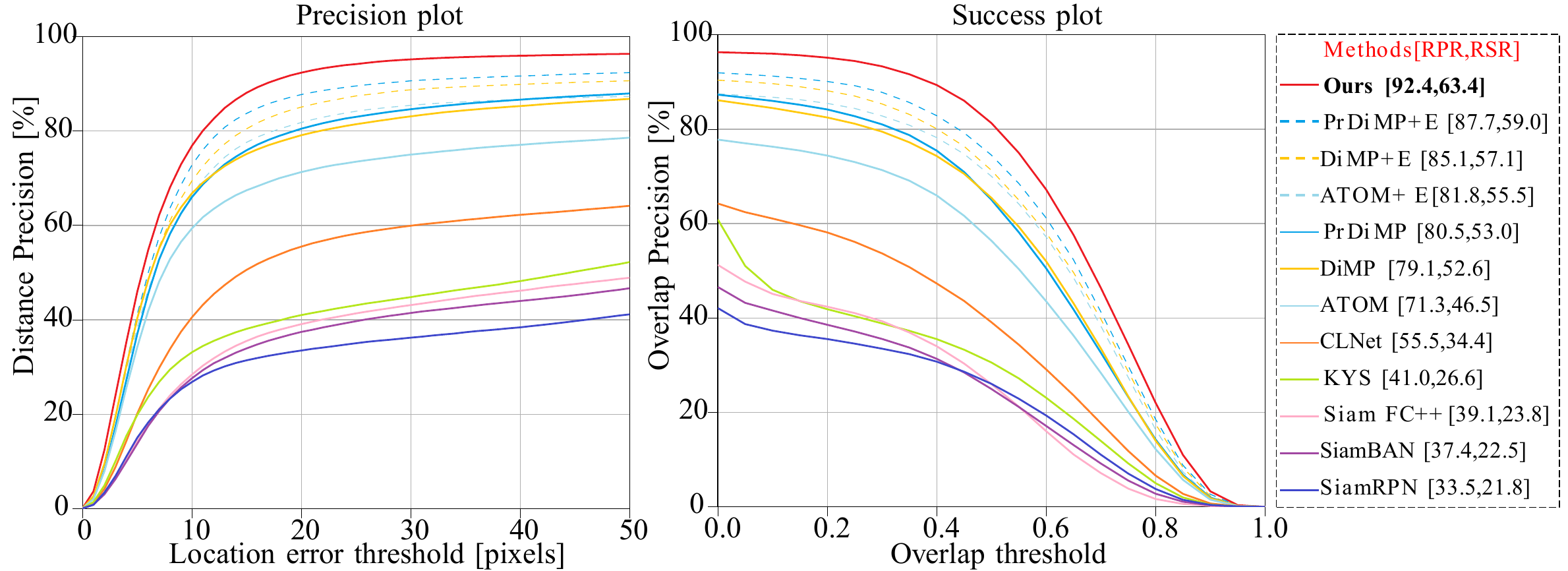}  \\

	\end{tabular}
	\caption{\bd{Precision (left) and Success (right) plot on FE108. In terms of both metric, our approach outperforms the state-of-the-art by a large margin.}}
	\label{fig:PRSR}
	\vspace{-0.5cm}
\end{figure}

\bd{Illustrated as the solid curves in Figure~\ref{fig:PRSR}, on FE108 dataset, our method outperforms other compared approaches by a large margin in terms of both precision and success rate. Specifically, the proposed approach achieves a 92.4\% overall RPR and 63.4\% RSR, and it outperforms the runner-up by 11.9\% and 10.4\%, respectively. To get more insights into the effectiveness of the proposed approach, we also show the performances under four different challenging conditions provided by FE108. As shown in Table~\ref{tab:attribute}, our method offers the best results under all four conditions, especially in LL and HDR conditions. Eight visual examples under different degraded conditions are shown in Figure~\ref{visual}, where we can see our approach can accurately track the target under all conditions.}

\def\wdenoising{1.0\linewidth}
\def\hdenoising{1.2in}
\begin{figure}[b]
    \vspace{-0.35cm}
	\setlength{\tabcolsep}{1.0pt}
	\centering
	\small
	\begin{tabular}{c}
		\includegraphics[width=\wdenoising]{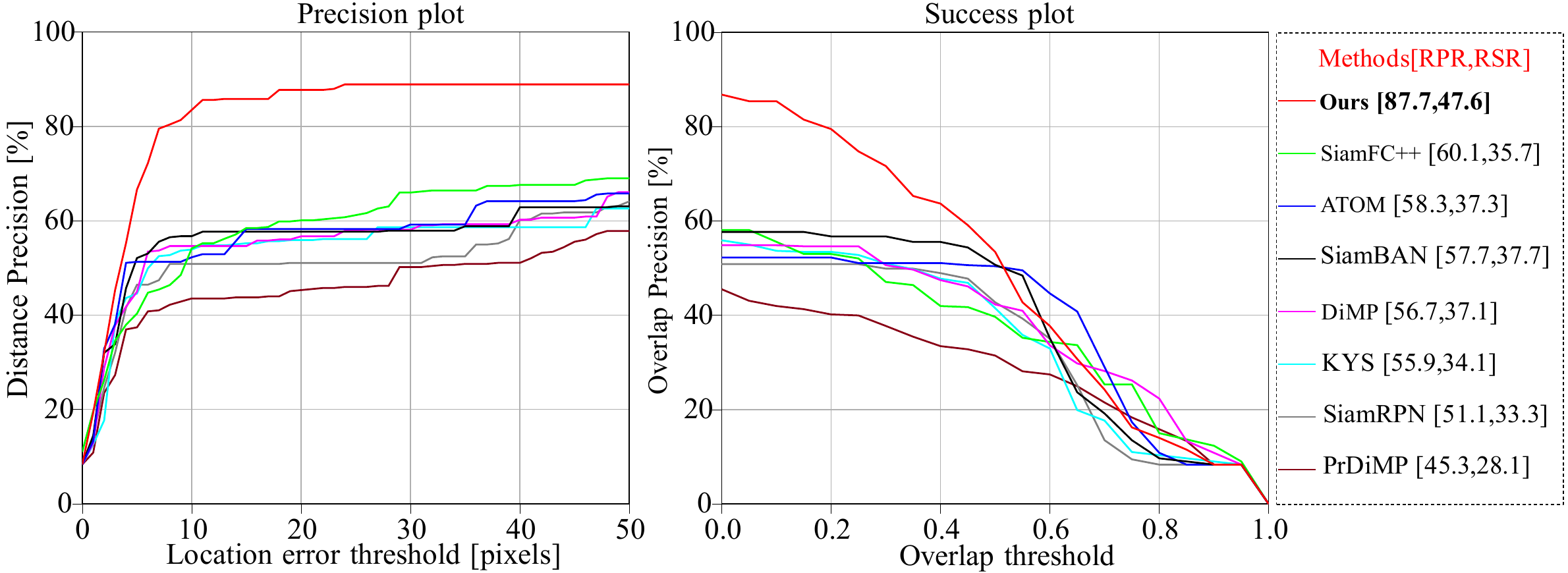}  \\ 
	\end{tabular}
	\caption{\bd{Precision (left) and Success (right) plot on EED~\cite{mitrokhin2018event}.}}
	\label{fig:PRSR_eed}
\end{figure}

\setlength{\tabcolsep}{0.85pt}
\begin{table}[t]
	\small

	    \scalebox{0.9}{
	\begin{tabular}{l|cc|cc|cc|cc|cc|cc}
		
		\hline   
		\hline
		\multirow{2}{*}{Methods} &  \multicolumn{2}{c|}{FD} &  \multicolumn{2}{c|}{LV}  & \multicolumn{2}{c|}{Occ} & \multicolumn{2}{c|}{WiB}
		& \multicolumn{2}{c|}{MO} & \multicolumn{2}{c}{ALL}\\
		\cline{2-13} 
		
		& \footnotesize{RSR} &\footnotesize{RPR} &\footnotesize{RSR} &\footnotesize{RPR} &\footnotesize{RSR} &\footnotesize{RPR} &\footnotesize{RSR} &\footnotesize{RPR} &\footnotesize{RSR} &\footnotesize{RPR} &\footnotesize{RSR} &\footnotesize{RPR} \\
		\hline                  
		SiamRPN~\cite{li2018high} &23  &43 &11 &10 &38 &40 &43 &63 &53 &100  &33  &51 \\
		ATOM~\cite{danelljan2019atom} &12   &19  &7 &12  &47 &60  &74 &100  &47  &100  &37  &58  \\
		DiMP~\cite{bhat2019learning} & 9  &19  &2 &4  &48 &60  &\textbf{79} &100  &50  &100  &37  &57 \\
		SiamFC++~\cite{xu2020siamfc++} & 17 &52 &10 &26 &45 &60 &58 &63 &50 &100 &36 &60 \\ %
		SiamBAN~\cite{chen2020siamese} & 22  &43 &8  &6 &36  &40  &69 &100  &54 &100  &38 &58 \\ %
		KYS~\cite{bhat2020know} &19  &38 &6 &19 &46 &60 &46 &63 &54 &100 &34 &56 \\
		CLNet~\cite{dongclnet} &10  &19 &2 &6 &19 &20 &13 &25 &4 &13 &9 &17 \\
		PrDiMP~\cite{danelljan2020probabilistic} &9  &14 &4 &22 &19 &20 &78 &100 &31 &70 & 28 & 45\\
	    Ours &\textbf{32}  &\textbf{81} &\textbf{35} &\textbf{98} & \textbf{48}  &\textbf{60} &69 &\textbf{100} &\textbf{55} & \textbf{100} & \textbf{48} & \textbf{88} \\	
		\hline
		\hline
	\end{tabular}}
	\caption{\bd{State-of-the-art comparison on EED~\cite{mitrokhin2018event} in terms of RSR and RPR.}} 
	\label{tab:eed}
	\vspace{-0.5cm}
\end{table}

\bd{
Even though EED~\cite{mitrokhin2018event} has very limited frames and associated events, it provides five challenging sequences: fast drone (FD), light variations (LV), occlusions (Occ), what is background (WiB), and multiple objects (MO). The first two sequences both record a fast moving drone under low illumination. The third and the fourth sequences record a moving ball with another object and a net as foreground, respectively.
The fifth sequence consists of multiple moving objects under normal lighting conditions. Therefore, we also compare our approach against other methods on EED~\cite{mitrokhin2018event}. The experimental results are shown in Figure~\ref{fig:PRSR_eed} and Table~\ref{tab:eed}. Our method significantly outperforms other approaches in all conditions except WiB. But with limited frames, the experimental result is less convincing and meaningful compared to the ones obtained from FE108.  
}

\bd{One question in our mind is whether combining the frame and event information can make other frame-based approaches outperform our approach. To answer this question, we combine APS and event aggregated frame by concatenation manner to train and test the top three frame-based performers (\ie, PrDiMP~\cite{danelljan2020probabilistic}, DiMP~\cite{bhat2019learning}, and ATOM~\cite{danelljan2019atom}). We report their RSR and RPR in Table~\ref{tab:attribute} and show the corresponding results as the dashed curve in Figure~\ref{fig:PRSR}. As we can see, our approach still outperforms all others by a considerable margin. It reflects the effectiveness of our specially designed cross-domain feature integrator. We also witness that the performance of the three chosen approaches can be improved significantly only by naively combining the frame and event domains. It means event information definitely plays an important role in dealing with degraded conditions.}

\def\wdenoising{0.9\linewidth}
\def\hdenoising{2.5in}
\begin{figure*}[]
	\setlength{\tabcolsep}{1.0pt}
	\centering
	\begin{tabular}{c}
		\includegraphics[width=\wdenoising,height=\hdenoising]{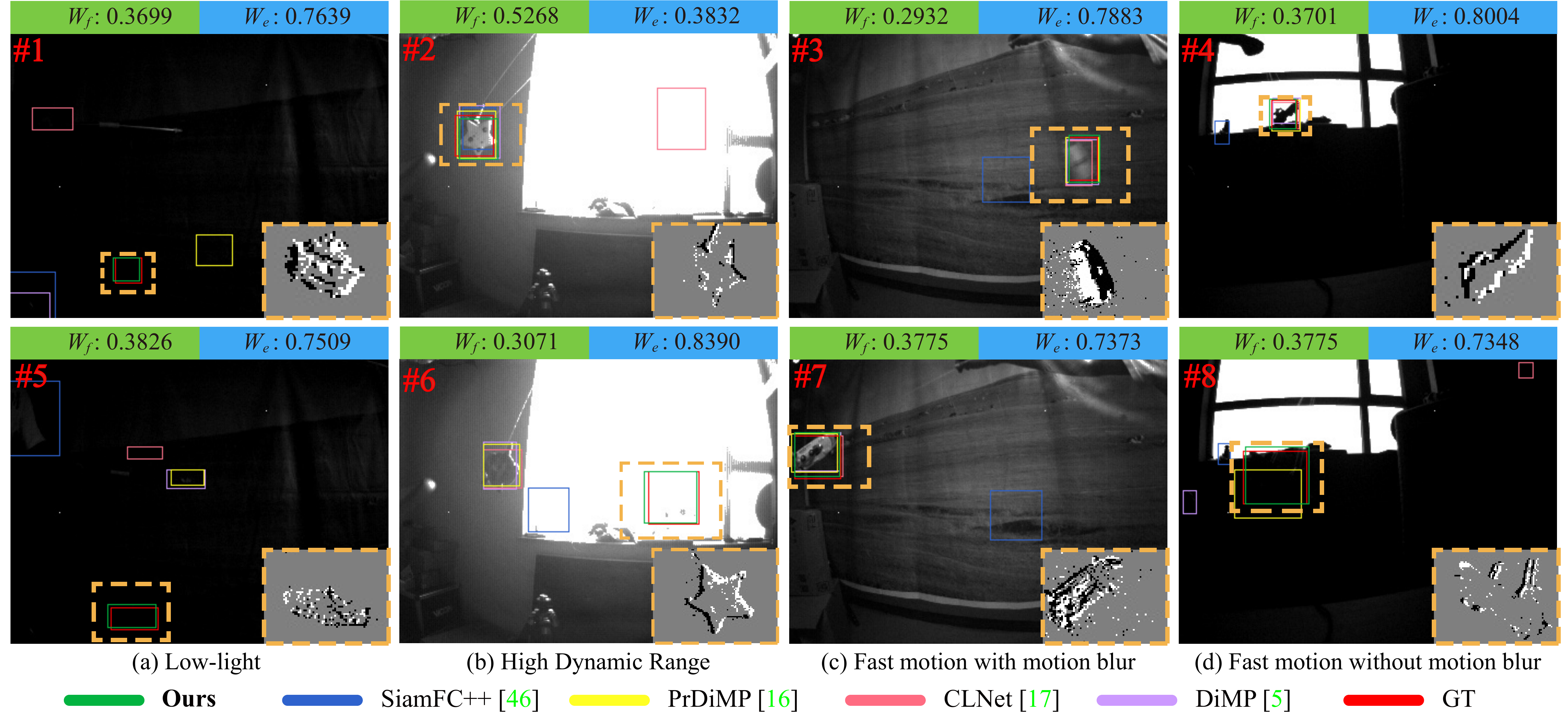}\\
		
	\end{tabular}
	\caption{\bd{Visual outputs of state-of-the-art algorithms on FE108 dataset. The lower-right dashed boxes show accumulated event frame of the dashed boxes inside the frames.}}
	\label{visual}
	\vspace{-0.4cm}
\end{figure*}

\subsection{Ablation Study}
\label{section:ablation}
\noindent
\tb{Multi-modal Input.}
\bd{We design the following experiments to show the effectiveness of multi-modal input.  1. Frame only: only using frames and \textbf{FFE}; 2. Event only: only using events and \textbf{EFE}; 3. Event to Frame: combining frames and events by concatenation as input to \textbf{FFE}; 4. Frame to Event: the same as 3, but input to \textbf{EFE}. For each setup, we train a dedicated model and test with it. As shown in the row \textit{A-D} of Table~\ref{tab:ablation}, the models with multi-modal inputs perform better than the ones with unimodal input. It 
shows the effectiveness of multi-modal fusion and our CDFI.}

\noindent

\setlength{\tabcolsep}{4.0pt}
\begin{table}[t]
	\centering

    \scalebox{0.9}{
	\begin{tabular}{cl|cccc}
		\hline
		\hline
		& Models & RSR$~\uparrow$ &OP$_{0.50}$$~\uparrow$ &OP$_{0.75}$$~\uparrow$  &RPR$~\uparrow$   \\
		
		\hline
	\textit{A.}	& Frame Only  & 45.6  &54.6   & 21.0  &73.1  \\
	\textit{B.}	& Event Only  & 52.0  & 63.2  & 20.3  &82.0   \\
	\textit{C.}	& Event to Frame &55.5   &70.0 &27.4   &82.8     \\
	\textit{D.}	& Frame to Event &53.6    &66.5  &25.9    &80.4     \\
		\hline
	
	\textit{E.}	& w/o EAB & 60.7  & 77.9 & 31.7  & 88.6    \\
	\textit{F.}	& w/o CDMS &  59.8 & 75.8   & 31.0  &88.1  \\
	\textit{G.}	& CDMS w/o SA  & 62.6  &79.8 &33.8   &91.5   \\
	\textit{H.}	& CDMS w/o CA  & 61.9  &78.8  &33.0   &90.7  \\
	\textit{I.}	& CDMS w/o AW  & 60.9  &77.2 &32.0   &89.9   \\
		
	\hline
	\textit{J.}	& TSLTD~\cite{chen2020end} &60.4  & 77.0 &31.2  &89.2   \\
	
	\textit{K.}	& Time Surfaces~\cite{lagorce2016hots}  & 61.4 &78.5   &32.9 &90.1   \\
	\textit{L.}	& Event Count~\cite{maqueda2018event}  &59.6  & 76.4   &27.4  & 88.6   \\
	\textit{M.}	& Event Frame~\cite{rebecq2017real}  &59.0  &74.5  &29.9  &87.7   \\
	\textit{N.}	& Zhu \textit{et al}.~\cite{zhu2019unsupervised} & 61.9 &79.2& 32.3 &  91.2   \\
	
	\hline
	\textit{O.}	& All $w=1$ & 61.3  &78.1 &31.6   &90.1     \\
	\hline
	\textit{P.}	& Ours  &\textbf{63.4}   &\textbf{81.3} &\textbf{34.4}   &\textbf{92.4}    \\
		\hline
		\hline
	\end{tabular}
	}
	\caption{\bd{Ablation study results.}}
	\label{tab:ablation}
	\vspace{-0.6cm}
\end{table}

\noindent
\bd{
\tb{Effectiveness of the proposed key components.} There are two key components in our approach: EAB and CDMS. Inside the CDMS, there are three primary schemes: self-attention (Eq.~\ref{eq:CDMS_SA}), cross-attention (Eq.~\ref{eq:CDMS_CA}), and adaptive weighting (Eq.~\ref{eq:CDMS_aw}). To verify their effectiveness, we modify the original model by dropping each of the components and retrain the modified models. Correspondingly, we obtain five retrained models: (i) without EAB; (ii) without CDMS; Inside CDMS, (iii) without self-attention (CDMS w/o SA); (iv) without cross-attention (CDMS w/o CA); (v) without adaptive weighting (CDMS w/o AW). The results of the five modified models are shown in the row \textit{E-I} of Table~\ref{tab:ablation}, respectively. Compared to the original model, removing CDMS has the most considerable impact on the performance, whereas removing the self-attention influences the least. It confirms the proposed CDMS is the key to our outstanding performance. Moreover, removing EAB also influences performance significantly. It shows that the EAB indeed enhances the extracted edge features.}

\bd{Inside CDMS, removing adaptive weighting scheme degrades performance the most. To get more insights into it, we report the estimated two weights (\ie, $w_f$ for the frame domain; $w_e$ for the event domain) of all eight visual examples in Figure~\ref{visual}. Except for the second one, the frame domain cannot provide reliable visual cues. Correspondingly, we can see the $w_e$ in these seven examples are significantly higher than $w_f$, whereas $w_e$ is much lower than $w_f$ in the second scene. The fourth one provides an interesting observation. We can see the object clearly in the frame domain, but $w_e$ is still higher than $w_f$. We think it is because the model is trained to focus on texture cues in the frame domain, but no texture cues can be extracted in this case. 
It is worthwhile to mention that only our method can successfully track the target in all examples.}

\noindent
\tb{Event Aggregation.}
\bd{For the events captured between two adjacent frames, we slice them into $n$ chunks in the time domain and then aggregate them as EFE's inputs. Here, we study the impacts of hyperparameter $n$. As shown in Table~\ref{tb:n_test}, both RSR and RPR scores increase with a larger $n$ value. However, with a larger $n$ value, it slows down the inference time. We can see $n=3$ offers the best trade-off between accuracy and efficiency. The way of aggregating events is another factor that has an impact on the performance. We conducted experiments with five commonly used event aggregation methods~\cite{chen2020end,lagorce2016hots,maqueda2018event,rebecq2017real,zhu2019unsupervised}. The results are shown in the row \textit{J-N} of Table~\ref{tab:ablation}, and our method still delivers the best performance. It suggests that discretizing the time dimension and leveraging the recent timpstamp information are effective for tracking. Another component associated with event aggregation is the weights in Eq.~\ref{equ:weight}, which are learned during the training process. 
We manually set the weights to 1 with $n=3$. The result is shown in row \textit{O} of Table~\ref{tab:ablation}, and we can see the corresponding performance is worse than the original model. 
}

\setlength{\tabcolsep}{10.0pt}
\begin{table}[t]
	\small
	\centering
	
	    \scalebox{0.9}{
	\begin{tabular}{c|cccccc}
		\hline
		\hline
		$n$ & 1  & 2 &3 &4 &5  & 6    \\
		\hline
		RPR$~\uparrow$  & 89.3  &90.1 &92.4 &92.6 &92.6   &92.7      \\
		RSR$~\uparrow$  & 60.2  & 61.7 & 63.4  &63.8  &63.4 &63.9      \\
		FPS$~\downarrow$ &35.1   &32.7  &30.1 &27.9   &25.2 &22.7      \\
		\hline
		\hline
	\end{tabular}}
	\caption{\bd{Trade-off between accuracy and efficiency introduced by the number of slices of event aggregation (\ie, $n$).}}
	\vspace{-0.46cm}
	\label{tb:n_test}
\end{table}

\section{Discussion and Conclusion}

\bd{
In this paper, we introduce a frame-event fusion-based approach for single object tracking. Our novel designed attention schemes effectively fuse the information obtained from both the frame and event domains. Besides, the novel developed weighting scheme is able to balance the contributions of the two domains adaptively. To enable further research on multi-modal learning and object tracking with events, we construct a large-scale dataset, FE108, comprising events, frames, and high-frequency annotations. 
Our approach outperforms frame-based state-of-the-art methods, which indicates leveraging the complementarity of events and frames boosts the robustness of object tracking in degraded conditions. Our current focus is on developing a cross-domain fusion scheme that can enhance visual tracking robustness, especially in degraded conditions. However, we have not leveraged the high measurement rate of event-based cameras to achieve low-latency tracking and the frame rate in the frame-domain bounds the tracking frequency of the proposed approach. One limitation of our frame-event-based dataset, FE108, is that no sequence contains the scenario of no events.  Our further work will focus on these two aspects: 1) We will investigate the feasibility of increasing tracking frequency by leveraging the high measurement rate of event-based cameras; 2) We will expand the FE108 by collecting more challenging sequences, especially with no events and more realistic scenes. 
}

~\\
\noindent
\jq{\textbf{Acknowledgements.} This work was supported in part by the National Natural Science Foundation of China under Grant 61632006, Grant 61972067, and the Innovation Technology Funding of Dalian (Project No. 2018J11CY010, 2020JJ26GX036).}

\clearpage

{\small
\bibliographystyle{ieee_fullname}
\bibliography{egbib}
}

\end{document}